\PassOptionsToPackage{table}{xcolor}

\documentclass[10pt]{article} 
\usepackage[preprint]{tmlr}

\usepackage{acro}
\usepackage{algorithm}
\usepackage{algpseudocodex}
\usepackage{amsmath}
\usepackage[british]{babel}
\usepackage{bookmark}
\usepackage{booktabs}
\usepackage{csquotes}
\usepackage[shortcuts]{extdash}
\usepackage{fontawesome5}
\usepackage{hyperref}
\usepackage{bookmark}
\usepackage{cleveref}
\usepackage{listings}
\usepackage{microtype}
\usepackage[all]{nowidow}
\usepackage{placeins}
\usepackage{pgfplots}
\usepackage{rotating}
\usepackage{siunitx}
\usepackage[most]{tcolorbox}
\usepackage{threeparttable}
\usepackage{tikz}
\usepackage{xcolor}
\usepackage{xurl}

\DeclareAcronym{aime}{
  long=American Invitational Mathematics Examination,
  short=\textsc{Aime},
}
\DeclareAcronym{dapo}{
  long=decoupled clip and dynamic sampling policy optimization,
  short=\textsc{dapo},
}
\DeclareAcronym{grpo}{
  long=group relative policy optimization,
  short=\textsc{grpo},
}
\DeclareAcronym{llm}{
  long=large language model,
  short=LLM,
}
\DeclareAcronym{lora}{
  long=low-rank adaptation,
  short=LoRA,
}
\DeclareAcronym{mada}{
  long=multi-agent debate-aware reinforcement learning,
  short=\textsc{mada-rl},
}
\DeclareAcronym{maporl}{
  long=multi-agent post-co-training for collaborative LLMs,
  short=MAPoRL,
}
\DeclareAcronym{marft}{
  long=multi-agent reinforcement fine-tuning,
  short=\textsc{marft},
}
\DeclareAcronym{oreo}{
  long=offline reasoning optimization,
  short=\textsc{oreo},
}
\DeclareAcronym{rl}{
  long=reinforcement learning,
  short=RL,
}
\DeclareAcronym{tina}{
  long=tiny reasoning models via LoRA,
  short=Tina,
}

\newtcblisting{prompt}{
  breakable,
  listing only,
  listing options={
    basicstyle=\ttfamily,
    breakatwhitespace=true,
    breakautoindent=false,
    breakindent=0pt,
    breaklines=true,
    columns=fullflexible,
    extendedchars=true,
    inputencoding=utf8,
    keepspaces=true,
    literate =
    {–}{{--}}1
    {—}{{---}}1
    {…}{{\ldots}}1
    {’}{{'}}1
    {‘}{{`}}1
    {”}{{''}}1
    {“}{{``}}1
    {±}{{$\pm$}}1
    {≤}{{$\le$}}1
    {≥}{{$\ge$}}1,
    postbreak=\mbox{\textcolor{gray}{$\hookrightarrow$}\space},
  }
}

\tikzset{
  box/.style={
    align=center,
    draw,
    font=\small,
    rounded corners,
  },
  critbox/.style={
    box,
    fill=my_gold,
    minimum width=2.1cm,
  },
  datasetbox/.style={
    box,
    fill=my_purple,
  },
  genbox/.style={
    box,
    fill=my_teal,
    minimum width=2.6cm,
  },
  inputbox/.style={
    box,
    fill=my_purple,
    minimum width=1.4cm,
  },
  mergebox/.style={
    box,
    fill=my_boxgray,
  },
  outbox/.style={
    box,
    fill=my_boxgray!50,
    minimum width=1.25cm,
  },
  subsetbox/.style={
    box,
    fill=my_purple!50,
    minimum width=2.3cm,
  },
}
\usetikzlibrary{
  calc,
  patterns,
  positioning,
  shapes.misc,
}

\definecolor{my_boxgray}{HTML}{CECECE}
\definecolor{my_darkgray}{HTML}{595959}
\definecolor{my_gold}{HTML}{F0C571}
\definecolor{my_gray}{HTML}{B8B8B8}
\definecolor{my_green}{HTML}{36B700}
\definecolor{my_lightgray}{HTML}{EDEDED}
\definecolor{my_purple}{HTML}{BB81BF}
\definecolor{my_red}{HTML}{E77173}
\definecolor{my_teal}{HTML}{59A89C}

\fontfamily{lmr}\selectfont
\DeclareFontShape{T1}{lmr}{b}{sc}{<->ssub*cmr/bx/sc}{}
\DeclareFontShape{T1}{lmr}{bx}{sc}{<->ssub*cmr/bx/sc}{}

\newcommand{\gray}[2]{\cellcolor{my_gray!#1!my_lightgray} #2}
\newcommand{\green}[2]{\cellcolor{my_green!#1!my_lightgray} #2}
\newcommand{\red}[2]{\cellcolor{my_red!#1!my_lightgray} #2}
\newcommand{\textlight}[1]{\textcolor{my_darkgray}{#1}}

\newlength{\sigcolw}

\title{MADA-RL: Multi-Agent Debate-Aware Reinforcement Learning for Parameter-Efficient Reasoning in Compact Models} 

\author{%
  \name{Martino M.\ L.\ Pulici} \email{martino.pulici@de.bosch.com} \\
  \addr{Bosch Center for Artificial Intelligence, Germany} \\
  \addr{LMU Munich, Germany}
  \AND{}%
  \name{Cuong Xuan Chu} \email{cuongxuan.chu@de.bosch.com} \\
  \addr{Bosch Center for Artificial Intelligence, Germany}
  \AND{}%
  \name{Evgeny Kharlamov} \email{evgeny.kharlamov@de.bosch.com} \\
  \addr{Bosch Center for Artificial Intelligence, Germany} \\
  \addr{University of Oslo, Norway}
  \AND{}%
  \name{Zifeng Ding} \email{zd320@cam.ac.uk} \\
  \addr{University of Cambridge, United Kingdom}
  \AND{}%
  \name{Volker Tresp} \email{volker.tresp@lmu.de} \\
  \addr{LMU Munich, Germany} \\
  \addr{Munich Center for Machine Learning, Germany}
  \AND{}%
  \name{Yunpu Ma} \email{cognitive.yunpu@gmail.com} \\
  \addr{LMU Munich, Germany} \\
  \addr{Munich Center for Machine Learning, Germany}
}

\def\month{MM} 
\def\year{YYYY} 
\def\openreview{\url{https://openreview.net/forum?id=XXXX}} 

\begin{document}

\maketitle

\begin{abstract}
  Large language models achieve strong reasoning performance, but often at prohibitive training cost\---a challenge that is especially acute for compact models ($\leq \qty{4}{B}$ parameters) trained under limited budgets.
  We introduce \textsc{mada-rl}, a post-training framework that specializes compact models into generator and critic roles and trains them with a debate-aware learning signal, fine-tuning only a small subset of parameters via LoRA adapters.
  Our central contribution is a counterfactual critic advantage: a dynamic, role-conditioned baseline that redefines the critic's advantage as its reward minus the generator ensemble's per-instance accuracy.
  This explicitly optimizes critics to improve over generator consensus rather than to merely reproduce a correct answer, yielding more targeted credit assignment than static mean-reward normalization.
  At deployment, the specialized agents are composed in a lightweight multi-round protocol.
  Across five mathematical reasoning benchmarks, \textsc{mada-rl} raises the accuracy of the DeepSeek-R1-Distill-Qwen-1.5B model from \qty{39.9}{\percent} to \qty{41.9}{\percent} (\num{+2.0} points, $p < \num{0.001}$) using \num{16} times fewer trainable parameters than fully fine-tuned baselines, placing it on the accuracy--trainable-parameter Pareto front. 
  It approaches, but does not surpass, the strongest baselines (DeepScaleR, \textsc{Still-3}), which are trained on substantially larger datasets; we analyse this gap and the associated inference-time cost directly.
  A controlled study isolates the source of \textsc{mada-rl}'s gains: the counterfactual advantage produces the highest critic improvement rate of any model evaluated, indicating that trained critics learn to correct generator errors rather than to imitate them.
\end{abstract}

\section{Introduction}

\Acp{llm} have achieved impressive performance across diverse language tasks, including translation, dialogue, reading comprehension, question answering, and open-ended text generation \citep{openai2024gpt,geminiteam2025gemini,qwenteam2025qwen25,qwenteam2025qwen3}.
Yet, multi-step logical reasoning remains a core challenge, especially for resource-constrained models with fewer than \qty{4}{B} parameters \citep{xu2025towards}.
While increasing model scale often yields better reasoning \citep{brown2020language}, it is prohibitively expensive in both training and inference \citep{hoffmann2022empirical,arora2025training}.

To bridge this gap, two complementary strategies have recently emerged.
The first leverages \ac{rl} to fine-tune models for reasoning.
Notable systems include OpenAI's o-series \citep{openai2024learning,openai2025introducing} and DeepSeek-R1 \citep{deepseekai2025deepseek}, which use \ac{rl} policy optimization \citep{shao2024deepseekmath,mei2025real} and preference learning \citep{christiano2017deep,ouyang2022training} to align model behaviours.
Despite impressive gains, \ac{rl} approaches often incur high computational cost \citep{hu2023aligning,wang2025tina}, risk training instability and model collapse \citep{cui2025process,yuan2025whats}, and depend heavily on reward-design quality.

The second strategy comprises methods that increase computational cost per query, a phenomenon commonly referred to as test-time scaling.
Among them, chain-of-thought prompting \citep{wei2023chain}, self-consistency \citep{wang2022self}, and tree-of-thought search \citep{yao2023tree} provide explicit multi-step reasoning, sampling diversity, and lookahead search, respectively.
Reflection and self-refinement methods further allow a single model to critique and improve its own outputs \citep{madaan2023self}.
Multi-agent debate frameworks take this further: by pitting agents against one another, they promote answer robustness and reduce superficial or shortcut reasoning \citep{irving2018ai,du2024improving,subramaniam2025multiagent}.

Prior attempts to integrate \ac{rl} with test-time prompting or debate \citep{park2025maporl,subramaniam2025multiagent} often involve significant computational overhead or architectural complexity.
We instead study how to combine these strands in a lightweight pipeline, addressing three limitations of prior \ac{rl}-based reasoning methods: they often require full-model fine-tuning, suffer from unstable credit assignment, and under-utilize test-time deliberation.
To this end, we propose \textbf{\ac{mada}}, a lightweight post-training framework that combines structured multi-agent RL with debate, while fine-tuning only a small fraction of parameters via \acl{lora} (\acs{lora}; \citealp{hu2022lora}) and \acl{grpo} (\acs{grpo}; \citealp{shao2024deepseekmath}).
We also introduce counterfactual advantage to stabilize learning and sharpen credit assignment among debating agents.

We demonstrate our approach on DeepSeek-R1-Distill-Qwen-1.5B, a \qty{1.5}{B}-parameter, distillation-based model widely used in small-scale reasoning research \citep{chen2025empirical,dang2025reinforcement,luo2025deepscaler,wang2025tina}.
Beyond our own fine-tuned agents, we evaluate a collection of community-released variants, assessing performance under single-agent and multi-agent debate conditions.
We focus on standard mathematical reasoning benchmarks and report rigorous statistical metrics, enabling more robust comparisons.
We also provide a detailed analysis of the training and inference cost trade-offs involved, and we isolate the source of \ac{mada}'s gains to improved critic behaviour rather than to additional test-time deliberation alone.

To summarize, our contributions include:

\begin{itemize}
  \item a counterfactual critic advantage for role-specialized \ac{rl}, constituting a dynamic, role-conditioned baseline that rewards critics for improving over the generator ensemble's per-instance accuracy, sharpening credit assignment without value models, replay buffers, or external verifiers
  \item \ac{mada}, a parameter-efficient post-training method that applies this signal to compact \acp{llm}, together with a suite of LoRA-fine-tuned agents built on DeepSeek-R1-Distill-Qwen-1.5B
  \item a controlled analysis that traces \ac{mada}'s gains to learned corrective behaviour rather than to additional test-time deliberation alone
  \item a comprehensive empirical study of existing \qty{1.5}{B} fine-tunings under a common role-specialized protocol, together with an accounting of the training-parameter and inference-token \mbox{trade-offs involved}.
\end{itemize}

The remainder of the paper is organized as follows: \cref{sec:debate} introduces the test-time debate algorithm, \cref{sec:mada-rl} presents the \ac{mada} training framework and advantage design, \cref{sec:experiments} describes the experimental setup and results, \cref{sec:related-work} reviews related work, and~\cref{sec:conclusion} concludes.

\section{Test-time debate}\label{sec:debate}

To contextualize our training objective, we must first define the test-time debate algorithm that our agents will operate within.
Our \ac{rl} framework relies on this procedure to generate the counterfactual advantage used to train critic agents.
Following \citet{du2024improving} and \citet{subramaniam2025multiagent}, we define the framework as:

\begin{itemize}
  \item multi-agent, as multiple agents provide their answers in parallel
  \item multi-round, as agents debate in sequence using previous round answers
  \item multi-role, as agents that generate answers (generators) need different expertise from those who assess previous responses (critics).
\end{itemize}

We use the term \emph{debate} in the deliberative sense of \citet{du2024improving}\---multiple model instances exposing their answers to one another and revising over rounds\---rather than the adversarial, judge-arbitrated sense of \citet{irving2018ai} and \citet{liang2024encouraging}.
Concretely, our protocol is parallel generation followed by sequential critic revision conditioned on the other agents' outputs; we make no claim of explicit argumentation or persuasion between agents.
This deliberately simple interaction is what makes the protocol cheap to deploy on compact models, and our contribution lies not in the protocol itself but in the training signal that shapes how critics behave within it (\cref{sec:training-critics}).

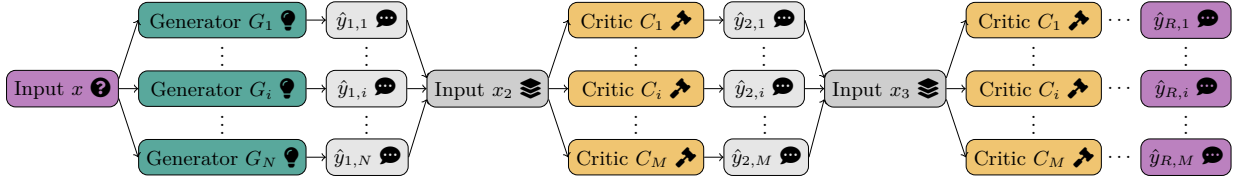
\begin{figure}[h]
  \centering
  \resizebox{\textwidth}{!}{
    \tikzset{node distance=5mm and 3mm}
    \begin{tikzpicture}
      \node [inputbox] (input) {Input $x$ \faQuestionCircle};
      \node [genbox] (model_n_1) [right=of input] {Generator $G_i$ \faLightbulb};
      \node [genbox] (model_1_1) [above=of model_n_1] {Generator $G_1$ \faLightbulb};
      \node [genbox] (model_N_1) [below=of model_n_1] {Generator $G_N$ \faLightbulb};
      \node [scale=0.9] at ($(model_1_1)!0.44!(model_n_1)$) {$\vdots$};
      \node [scale=0.9] at ($(model_n_1)!0.44!(model_N_1)$) {$\vdots$};
      \draw [->] (input.10) -- (model_1_1.west);
      \draw [->] (input) -- (model_n_1);
      \draw [->] (input.350) -- (model_N_1.west);
      \node [outbox] (y_1_n) [right=of model_n_1] {$\hat{y}_{1,i}$ \faCommentDots};
      \node [outbox] (y_1_1) [right=of model_1_1] {$\hat{y}_{1,1}$ \faCommentDots};
      \node [outbox] (y_1_N) [right=of model_N_1] {$\hat{y}_{1,N}$ \faCommentDots};
      \draw [->] (model_1_1) -- (y_1_1);
      \draw [->] (model_n_1) -- (y_1_n);
      \draw [->] (model_N_1) -- (y_1_N);
      \node [scale=0.9] at ($(y_1_1)!0.44!(y_1_n)$) {$\vdots$};
      \node [scale=0.9] at ($(y_1_n)!0.44!(y_1_N)$) {$\vdots$};
      \node [mergebox] (summarize) [right=of y_1_n] {Input $x_2$ \faLayerGroup};
      \draw [->] (y_1_1.east) -- (summarize.170);
      \draw [->] (y_1_n) -- (summarize);
      \draw [->] (y_1_N.east) -- (summarize.190);
      \node [critbox] (model_n_2) [right=of summarize] {Critic $C_i$ \faGavel};
      \node [critbox] (model_1_2) [above=of model_n_2] {Critic $C_1$ \faGavel};
      \node [critbox] (model_N_2) [below=of model_n_2] {Critic $C_M$ \faGavel};
      \node [scale=0.9] at ($(model_1_2)!0.44!(model_n_2)$) {$\vdots$};
      \node [scale=0.9] at ($(model_n_2)!0.44!(model_N_2)$) {$\vdots$};
      \draw [->] (summarize.10) -- (model_1_2.west);
      \draw [->] (summarize) -- (model_n_2);
      \draw [->] (summarize.350) -- (model_N_2.west);
      \node [outbox] (y_2_n) [right=of model_n_2] {$\hat{y}_{2,i}$ \faCommentDots};
      \node [outbox] (y_2_1) [right=of model_1_2] {$\hat{y}_{2,1}$ \faCommentDots};
      \node [outbox] (y_2_N) [right=of model_N_2] {$\hat{y}_{2,M}$ \faCommentDots};
      \draw [->] (model_1_2) -- (y_2_1);
      \draw [->] (model_n_2) -- (y_2_n);
      \draw [->] (model_N_2) -- (y_2_N);
      \node [scale=0.9] at ($(y_2_1)!0.44!(y_2_n)$) {$\vdots$};
      \node [scale=0.9] at ($(y_2_n)!0.44!(y_2_N)$) {$\vdots$};
      \node [mergebox] (summarize2) [right=of y_2_n] {Input $x_3$ \faLayerGroup};
      \draw [->] (y_2_1.east) -- (summarize2.170);
      \draw [->] (y_2_n) -- (summarize2);
      \draw [->] (y_2_N.east) -- (summarize2.190);
      \node [critbox] (model_n_3) [right=of summarize2] {Critic $C_i$ \faGavel};
      \node [critbox] (model_1_3) [above=of model_n_3] {Critic $C_1$ \faGavel};
      \node [critbox] (model_N_3) [below=of model_n_3] {Critic $C_M$ \faGavel};
      \node [scale=0.9] at ($(model_1_3)!0.44!(model_n_3)$) {$\vdots$};
      \node [scale=0.9] at ($(model_n_3)!0.44!(model_N_3)$) {$\vdots$};
      \draw [->] (summarize2.10) -- (model_1_3.west);
      \draw [->] (summarize2) -- (model_n_3);
      \draw [->] (summarize2.350) -- (model_N_3.west);
      \tikzset{node distance=6mm}
      \node [inputbox] (y_M_n) [right=of model_n_3] {$\hat{y}_{R,i}$ \faCommentDots};
      \node [inputbox] (y_M_1) [right=of model_1_3] {$\hat{y}_{R,1}$ \faCommentDots};
      \node [inputbox] (y_M_N) [right=of model_N_3] {$\hat{y}_{R,M}$ \faCommentDots};
      \path (model_n_3) -- (y_M_n) node [midway, scale=0.9] {$\cdots$};
      \path (model_1_3) -- (y_M_1) node [midway, scale=0.9] {$\cdots$};
      \path (model_N_3) -- (y_M_N) node [midway, scale=0.9] {$\cdots$};
      \node [scale=0.9] at ($(y_M_1)!0.44!(y_M_n)$) {$\vdots$};
      \node [scale=0.9] at ($(y_M_n)!0.44!(y_M_N)$) {$\vdots$};
    \end{tikzpicture}
  }
  \caption{
    Multi-agent debate, adapted from \citet{subramaniam2025multiagent}.
  }\label{fig:debate}
\end{figure}

\Cref{fig:debate,alg:debate} detail the procedure: during the first debate round, each generator agent $G_i$ generates an answer to a question $x$; then, for all successive debate rounds, each critic agent $C_i$ produces an updated guess based on the original question and concatenated answers from the previous round; at the end, accuracy is computed using only answers from the final debate round.

\begin{algorithm}
  \caption{Multi-agent debate}\label{alg:debate}
  \textbf{Input}: Model $A$, $N$ generator LoRA adapters $\{L_{G, i}\}$, $M$ critic LoRA adapters $\{L_{C, i}\}$, dataset $\mathcal{D} = \{(x_i, y_i)\}$ \\
  \textbf{Parameter}: Number of debate rounds $R$ \\
  \textbf{Output}: Final round accuracy
  \begin{algorithmic}[1]
    \State{$\texttt{success} \gets 0$} \Comment{Initialize success counter}
    \State{${\{G_i\}}_{i=1}^N \gets {\{L_{G, i}(A)\}}_{i=1}^N$} \Comment{Instantiate generator agents}
    \State{${\{C_i\}}_{i=1}^M \gets {\{L_{C, i}(A)\}}_{i=1}^M$} \Comment{Instantiate critic agents}
    \Statex{}
    \ForAll{$(x, y) \in \mathcal{D}$}
    \State{${\{y _{1, i}\}}_{i=1}^N \gets {\{G_i(x)\}}_{i=1}^N$} \Comment{Generate generators' responses}
    \Statex{}
    \For{$r \in \{2, \dots, R\}$}
    \State{$x_r \gets [x; \{y_{r-1, i}\}]$} \Comment{Concatenate question and previous round responses}
    \State{${\{y_{r, i}\}}_{i=1}^M \gets {\{C_i(x_r)\}}_{i=1}^M$} \Comment{Generate critics' responses}
    \EndFor{}
    \Statex{}
    \State{$\texttt{success} \gets \texttt{success} + \sum_{i=1}^{M} \mathbf{1}(y_{R, i} \equiv y)$} \Comment{Update success counter}
    \EndFor{}
    \Statex{}
    \State{$\texttt{accuracy} \gets \frac{\texttt{success}}{M |\mathcal{D}|}$} \Comment{Compute accuracy}
    \State{\textbf{return} \texttt{accuracy}}
  \end{algorithmic}
\end{algorithm}

While we adopt the role specialization of agents into generators and critics from \citet{subramaniam2025multiagent}, we intentionally omit another enhancement proposed in the same work: replacing response concatenation between debate rounds with summarization.
This choice reflects our aim to keep the method simple and memory efficient.
Although summarization has been shown to yield modest performance improvements, it would require an additional model and introduce non-trivial computational and memory overhead.
Given that our setting involves only a few agents, whose responses comfortably fit within the model's context window, we found this trade-off unnecessary.
However, in scenarios with a larger number of agents, incorporating a summarizer could represent a valuable extension.

\section{Multi-agent debate-aware reinforcement learning}\label{sec:mada-rl}

In this section, we introduce \ac{mada}, our lightweight \ac{rl} framework that employs multiple agents with distinct roles: generators and critics.
The training process follows a two-stage design, where each stage uses the same underlying RL algorithm but employs different datasets and advantage computations, tailored to the specific objectives of the two agent types.
We begin by outlining the overall framework, then describe the training procedures for generator and critic agents in detail.

\subsection{The MADA-RL framework}
Our framework builds on \ac{grpo} \citep{shao2024deepseekmath}, a value-model-free policy optimization method that has been effectively employed to train the DeepSeek-R1 model family \citep{deepseekai2025deepseek}.
Inspired by \citet{subramaniam2025multiagent}, we train each agent on a disjoint subset of the data to foster role specialization, an approach shown to enhance multi-agent debate outcomes.
To maintain efficiency and scalability, \ac{mada} uses LoRA adapters \citep{hu2022lora}, allowing fine-tuning of only a small fraction of the model parameters while keeping the base weights frozen.
This significantly reduces computational and memory overhead, making our method accessible even in resource-constrained environments.

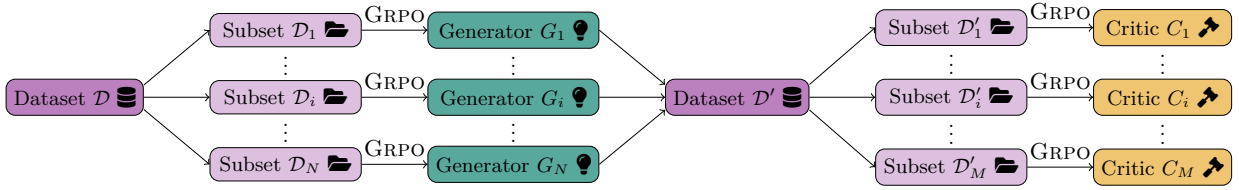
\begin{figure}[h]
  \centering
  \resizebox{\textwidth}{!}{
    \tikzset{node distance=5mm and 10mm}
    \begin{tikzpicture}
      \node [datasetbox] (input) {Dataset $\mathcal{D}$ \faDatabase};
      \node [subsetbox] (model_n_1) [right=of input] {Subset $\mathcal{D}_i$ \faFolderOpen};
      \node [subsetbox] (model_1_1) [above=of model_n_1] {Subset $\mathcal{D}_1$ \faFolderOpen};
      \node [subsetbox] (model_N_1) [below=of model_n_1] {Subset $\mathcal{D}_N$ \faFolderOpen};
      \node [scale=0.9] at ($(model_1_1)!0.44!(model_n_1)$) {$\vdots$};
      \node [scale=0.9] at ($(model_n_1)!0.44!(model_N_1)$) {$\vdots$};
      \draw [->] (input.10) -- (model_1_1.west);
      \draw [->] (input) -- (model_n_1);
      \draw [->] (input.350) -- (model_N_1.west);
      \node [genbox] (y_1_n) [right=of model_n_1] {Generator $G_i$ \faLightbulb};
      \node [genbox] (y_1_1) [right=of model_1_1] {Generator $G_1$ \faLightbulb};
      \node [genbox] (y_1_N) [right=of model_N_1] {Generator $G_N$ \faLightbulb};
      \node [scale=0.9]at ($(y_1_1)!0.44!(y_1_n)$) {$\vdots$};
      \node [scale=0.9]at ($(y_1_n)!0.44!(y_1_N)$) {$\vdots$};
      \draw [->] (model_1_1) -- node [midway, above] {\Acs{grpo}} (y_1_1);
      \draw [->] (model_n_1) -- node [midway, above] {\Acs{grpo}} (y_1_n);
      \draw [->] (model_N_1) -- node [midway, above] {\Acs{grpo}} (y_1_N);
      \node [datasetbox] (summarize) [right=of y_1_n] {Dataset $\mathcal{D}'$ \faDatabase};
      \draw [->] (y_1_1.east) -- (summarize.170);
      \draw [->] (y_1_n) -- (summarize);
      \draw [->] (y_1_N.east) -- (summarize.190);
      \node [subsetbox] (model_n_2) [right=of summarize] {Subset $\mathcal{D}'_i$ \faFolderOpen};
      \node [subsetbox] (model_1_2) [above=of model_n_2] {Subset $\mathcal{D}'_1$ \faFolderOpen};
      \node [subsetbox] (model_N_2) [below=of model_n_2] {Subset $\mathcal{D}'_M$ \faFolderOpen};
      \node [scale=0.9]at ($(model_1_2)!0.44!(model_n_2)$) {$\vdots$};
      \node [scale=0.9]at ($(model_n_2)!0.44!(model_N_2)$) {$\vdots$};
      \draw [->] (summarize.10) -- (model_1_2.west);
      \draw [->] (summarize) -- (model_n_2);
      \draw [->] (summarize.350) -- (model_N_2.west);
      \node [critbox] (y_M_n) [right=of model_n_2] {Critic $C_i$ \faGavel};
      \node [critbox] (y_M_1) [right=of model_1_2] {Critic $C_1$ \faGavel};
      \node [critbox] (y_M_N) [right=of model_N_2] {Critic $C_M$ \faGavel};
      \node [scale=0.9]at ($(y_M_1)!0.44!(y_M_n)$) {$\vdots$};
      \node [scale=0.9]at ($(y_M_n)!0.44!(y_M_N)$) {$\vdots$};
      \draw [->] (model_1_2) -- node [midway, above] {\Acs{grpo}} (y_M_1);
      \draw [->] (model_n_2) -- node [midway, above] {\Acs{grpo}} (y_M_n);
      \draw [->] (model_N_2) -- node [midway, above] {\Acs{grpo}} (y_M_N);
    \end{tikzpicture}
  }
  \caption{
    The \ac{mada} training pipeline.
  }\label{fig:mada-rl}
\end{figure}

\Cref{fig:mada-rl} presents the complete \ac{mada} training pipeline, and~\cref{alg:mada-rl} goes into further details: the first block (lines 1--5) trains the generator agents; the second block (lines 6--12) constructs the debate-aware dataset $\mathcal{D}'$ used to train the critic agents; the final block (lines 13--7) mirrors the generator training loop, but applies it to the critic agents.

\begin{algorithm}
  \caption{\Acs{mada} training}\label{alg:mada-rl}
  \textbf{Input}: Model $A$, LoRA adapter $L$, dataset $\mathcal{D} = \{(x_i, y_i)\}$ \\
  \textbf{Parameters}: Number of generators $N$, number of critics $M$ \\
  \textbf{Output}: Fine-tuned LoRA adapters ${\{L_{G, i}\}}_{i=1}^N$ and ${\{L_{C, i}\}}_{i=1}^M$
  \begin{algorithmic}[1]
    \State{${\{\mathcal{D}_i\}}_{i=1}^N \gets \mathcal{D}$} \Comment{Partition dataset into disjoint subsets}
    \For{$i \in \{1, \ldots, N\}$}
    \State{$G_i \gets L(A)$} \Comment{Instantiate generator from base model and blank LoRA adapter}
    \State{\textbf{train} $G_i$ on $\mathcal{D}_i$} \Comment{Train generator with \acs{grpo} advantage}
    \State{\textbf{save} $L_{G, i}$} \Comment{Save the best LoRA adapter}
    \EndFor{}
    \Statex{}
    \State{${\{G_i\}}_{i=1}^N \gets {\{L_{G, i}(A)\}}_{i=1}^N$} \Comment{Instantiate generator agents}
    \State{$\mathcal{D}' \gets \emptyset$} \Comment{Initialize empty critic dataset}
    \ForAll{$(x, y) \in \mathcal{D}$}
    \State{${\{\hat{y}_i\}}_{i=1}^N \gets {\{G_i(x)\}}_{i=1}^N$} \Comment{Generate generators' responses}
    \State{$x' \gets \left[x ; {\{\hat{y}_i\}}_{i=1}^N\right]$} \Comment{Concatenate question and generators' responses}
    \State{$\mathrm{acc}_G \gets \operatorname{mean}\left({\{\mathbf{1}(\hat{y}_i \equiv y)\}}_{i=1}^N\right)$} \Comment{Compute mean generator accuracy}
    \State{$\mathcal{D}' \gets \mathcal{D}' \cup \{(x', y, \mathrm{acc}_G)\}$} \Comment{Update critic dataset}
    \EndFor{}
    \Statex{}
    \State{${\{\mathcal{D}'_i\}}_{i = 1}^M \gets \mathcal{D}'$} \Comment{Partition critic dataset into disjoint subsets}
    \For{$i \in \{1, \ldots, M\}$}
    \State{$C_i \gets L(A)$} \Comment{Instantiate critic from base model and blank LoRA adapter}
    \State{\textbf{train} $C_i$ on $\mathcal{D}'_i$} \Comment{Train critic with counterfactual advantage}
    \State{\textbf{save} $L_{C, i}$} \Comment{Save the best LoRA adapter}
    \EndFor{}
    \Statex{}
    \State{\textbf{return} ${\{L_{G, i}\}}_{i=1}^N$ and ${\{L_{C, i}\}}_{i=1}^M$} \Comment{Return generators' and critics' LoRA adapters}
  \end{algorithmic}
\end{algorithm}

\subsection{Training generator agents}\label{sec:training-generators}

Generator agents are trained independently as single-agent models.
This setup is appropriate because they are only responsible for producing the initial responses during inference and do not participate in or receive feedback from subsequent debate rounds.
Treating them as isolated agents simplifies the training process and reflects their role in the system architecture.
A similar strategy is adopted by \citet{subramaniam2025multiagent}, who report that independently trained generators can effectively support multi-agent reasoning when paired with collaborative or evaluative agents.

To incentivize high-quality outputs, each generator is trained using a composite reward that balances two core aspects of answer quality: correctness and brevity.
Specifically, the reward function is a weighted sum of an accuracy reward and a length reward, with a 2:1 weighting in favor of accuracy.
This prioritization reflects the primary goal of producing correct answers, while also encouraging conciseness to avoid verbosity.
This design is informed by prior \ac{rl} studies such as those by \citet{chen2025empirical} and \citet{wang2025tina}, which demonstrate the benefits of multi-objective reward shaping in language model fine-tuning.

\paragraph{Accuracy reward.}
Given that our datasets consist of mathematically grounded problems, we define the reward signal as the binary function
\begin{equation*}
  R_\mathrm{acc}(\hat{y}, y) = \mathbf{1}(\hat{y} \equiv y)
\end{equation*}
based on symbolic equivalence between the model's prediction $\hat{y}$ and the ground truth answer $y$.
We use the equivalence symbol $\equiv$ instead of the equality symbol $=$ to reflect the method of evaluation: answers are parsed and normalized using a \LaTeX-to-SymPy converter \citep{kydlicek2025latex2sympy2} and verified for symbolic equality using a mathematical reasoning engine \citep{kydlicek2025math}.
This approach enables reliable reward computation even in the presence of minor formatting differences or expression reordering.

\paragraph{Length reward.}
As language models are prone to verbosity, especially during \ac{rl}, we introduce a length-based reward to promote concise outputs.
Following the method proposed by \citet{kimiteam2025kimi}, we penalize unnecessarily long completions based on token count, defining a length reward as
\begin{equation*}
  R_\mathrm{len}(\hat{y}, y) =
  \begin{cases}
    \lambda & \hat{y} \equiv y \\
    \min(0, \lambda) & \hat{y} \not\equiv y
  \end{cases}
\end{equation*}
\begin{equation*}
  \lambda = 0.5 - \frac{\operatorname{len}(\hat{y}) - \min_i \operatorname{len}(\hat{y}_i)}{\max_i \operatorname{len}(\hat{y}_i) - \min_i \operatorname{len}(\hat{y}_i)}
\end{equation*}
where $\operatorname{len}(\hat{y})$ is the number of tokens in the generated answer and $\{\hat{y}_i\}$ represents the set of sampled outputs for the same prompt within a single \ac{grpo} sampling group.
The length reward ranges from \num{-0.5} (for the longest correct output) to \num[retain-explicit-plus]{+0.5} (for the shortest correct output in the group), with incorrect outputs capped at zero to avoid reward hacking through trivially short but incorrect completions.

\paragraph{Final generator reward.}
The overall generator reward combines both objectives into a single scalar signal used for policy gradient updates, resulting in the final reward function
\begin{equation*}
  R_G(\hat{y}, y) = 2 \, R_\mathrm{acc}(\hat{y}, y) + R_\mathrm{len}(\hat{y}, y)
\end{equation*}
ensuring that accurate answers are strongly rewarded, while still giving a slight preference to brevity among correct outputs.
It also provides a safeguard against degenerate solutions, such as repetitive or vacuous~responses.

\subsection{Training critic agents}\label{sec:training-critics}

The critic training process reuses much of the generator training pipeline but incorporates two key modifications to account for the critic's distinct role.
First, the input dataset is enriched to include not only the original question but also the set of generator responses from the initial debate round.
Second, the policy update is guided by a counterfactual advantage signal, which sharpens credit assignment by measuring the critic's performance relative to its generator peers.

\paragraph{Counterfactual advantage.}
The critic's objective is not merely to produce a correct answer, but to do so especially in situations where the generators fail.
To formalize this, we define the advantage signal for the critic by first calculating the critic's total reward as
\begin{equation*}
  R_C(\hat{y}, y) = 2 \, R_\mathrm{acc}(\hat{y}, y) + R_\mathrm{len}(\hat{y}, y)
\end{equation*}
using the same composite function as the generators.
We then compute the advantage by subtracting a dynamic, role-conditioned baseline from this total reward.
This baseline is derived from the per-instance average generator accuracy $\mathrm{acc}_G$, computed at the time of the critic's dataset creation.
To ensure the baseline is scaled consistently with the primary component of the critic's reward, we multiply this average accuracy by two.
The derivation proceeds as
\begin{equation*}
  A_C(\hat{y}, y, \mathrm{acc}_G) = R_C(\hat{y}, y) - 2 \, \mathrm{acc}_G = 2 \, (R_\mathrm{acc}(\hat{y}, y) - \mathrm{acc}_G) + R_\mathrm{len}(\hat{y}, y)
\end{equation*}
providing a more informative learning signal.
The term in parentheses $(R_\mathrm{acc}(\hat{y}, y) - \mathrm{acc}_G)$ directly compares the critic's correctness to the generator average, and scaling it by two ensures this comparison is weighted appropriately.
A positive advantage strongly rewards the critic for outperforming the generator consensus on correctness, while a negative advantage penalizes it for underperforming.

\section{Experiments}\label{sec:experiments}

\subsection{Experimental setup}

For our main experiment, we trained a total of six agents\---three generators and three critics\---on random splits from the \textsc{Still-3}-Preview-RL-Data dataset \citep{rucaibox2025rl}, which contains \num{29925} mathematical reasoning problems.
To adapt the dataset to our multi-agent setting, we created three disjoint training subsets of \num{7500} problems each, along with a shared validation set of \num{750} problems for checkpoint selection.

All agents were fine-tuned starting from the DeepSeek-R1-Distill-Qwen-1.5B model \citep{deepseek2025distill}.
Training was performed using the \ac{grpo} algorithm \citep{shao2024deepseekmath}, in combination with the AdamW optimizer \citep{loshchilov2019decoupled} and \ac{lora} adapters \citep{hu2022lora}.
This setup allowed for efficient fine-tuning with a minimal memory footprint.
Training and inference hyperparameters are reported in the appendix, together with details about the computing infrastructure.

Simple accuracy is used for all experiments and all results are computed averaging the results of all agent instances.
In situations where only a subset of agents are used (as in single-agent, two-agent, and homogeneous settings of~\cref{sec:ablations}), multiple runs with all agent combinations are averaged.
To assess whether differences between models are meaningful, we compare per-seed average accuracies (across the five benchmarks, ten seeds per model) using Welch's two-sample $t$-test; results are reported in~\cref{tab:significance} and, in full pairwise form, in~\cref{sec:significance-full}.
Answer correctness was evaluated by parsing model generations and ground truth answers into symbolic form as described in~\cref{sec:training-generators}.

\subsubsection{Baselines}

We compare our agents against seven baselines built on the same base model, DeepSeek-R1-Distill-Qwen-1.5B, ensuring controlled and fair comparisons across different fine-tuning strategies.
These baselines are:

\begin{itemize}
  \item \textbf{DeepSeek-R1} \citep{deepseek2025distill}, a reasoning-oriented fine-tuning of Qwen2-1.5B \citep{qwen2024qwen2}, distilled from the flagship DeepSeek-R1 model \citep{deepseek2025deepseek} 
  \item \textbf{\textsc{Still-3}} \citep{rucaibox2025still}, inspired by slow-thinking paradigms such as o1 \citep{openai2024learning}, focusing on deliberate multi-step reasoning
  \item \textbf{DeepScaleR} \citep{agentica2025deepscaler}, which applies \ac{grpo} with an accuracy-based reward, emphasizing longer-context reasoning
  \item \textbf{Open-RS1}, \textbf{Open-RS2}, and \textbf{Open-RS3} \citep{knovelengineering2025rs1,knovelengineering2025rs2,knovelengineering2025rs3}; lightweight fine-tunings that prioritize low-resource adaptability while preserving reasoning performance
  \item \textbf{Tina} \citep{tina2025r1}, a \ac{lora}-based variant from the Tina family.
\end{itemize}

For a fairer comparison, we also re-trained the best-performing baselines (DeepScaleR and \textsc{Still-3}) using the original datasets and methods but with LoRA, yielding \textbf{DeepScaleR-LoRA} and \textbf{\textsc{Still-3}-LoRA}.

\subsubsection{Benchmarks}

We evaluate all models on five established benchmarks covering competition-style and academic mathematical~reasoning:

\begin{itemize}
  \item \textbf{\textsc{Math-500}} \citep{huggingface2024math}, a representative subset of the \textsc{Math} benchmark \citep{lightman2024lets}, featuring challenging competition-level problems across diverse topics
  \item \textbf{\textsc{Aime 2024}} and \textbf{\textsc{Aime 2025}} \citep{huggingface2025aime,lin2025aime}, consisting of full \ac{aime} problem sets, known for their structured, multi-step algebraic reasoning
  \item \textbf{AMC-23} \citep{knovelengineering2025amc}, including problems from the American Mathematics Competition, generally shorter than \ac{aime} but still requiring careful reasoning
  \item \textbf{Minerva-Math} \citep{knovelengineering2025minerva} contains university-level problems spanning mathematics and related \textsc{stem} fields, often involving symbolic manipulation and domain-specific modelling.
\end{itemize}

\subsection{Main results}

We applied the inference procedure of~\cref{sec:debate} across all benchmarks.
Our models (\textbf{\textsc{Mada}}) use the pipeline directly; for the single-agent baselines we instantiate identical copies in every generator and critic role, keeping the same agent count and rounds for a controlled comparison.
Mirroring the training setting, we use three generators, three critics, and two debate rounds.

\Cref{tab:main-results} reports average accuracy and standard error for all models, with the full pairwise Welch's $t$-tests in~\cref{sec:significance-full}.
The two models fine-tuned on the largest datasets (DeepScaleR and \textsc{Still-3}) clearly perform best, with DeepScaleR significantly ahead of \textsc{Still-3} ($p = \num{0.025}$); the other community fine-tunes (Open-RS1, Open-RS2, Open-RS3, Tina) are statistically indistinguishable from the base DeepSeek-R1 (all~$p > \num{0.20}$).

\begin{table}[t]
  \centering
  \sisetup{table-format=2.1(2)}
  \begin{threeparttable}
    \caption{Results of the main experiment, with accuracy values shown in base \num{100}}\label{tab:main-results}
    \begin{tabular}{l S[table-format=2.2(3)] *{5}S}
      \toprule
      Model                 & \textsc{Math-500}       & \textsc{Aime 2024}     & \textsc{Aime 2025}    & {AMC-23}              & {Minerva-Math}        & {Average} \\
      \midrule
      DeepSeek-R1           & 69.9(0.2)               & 23.0(1.0)              & 19.6(0.8)             & 65.3(0.9)             & 21.6(0.3)             & 39.9(0.3) \\
      \midrule
      \textsc{Still-3}      & \green{79}{72.5(0.2)}   & \green{94}{28.4(1.3)}  & 20.9(0.6)             & \green{67}{70.7(1.0)} & \green{34}{23.2(0.4)} & \green{66}{43.1(0.4)} \\
      DeepScaleR            & \green{88}{72.81(0.19)} & \green{100}{28.8(0.8)} & \green{54}{22.4(0.8)} & \green{76}{71.4(0.9)} & \green{92}{26.0(0.3)} & \green{90}{44.3(0.3)} \\
      Open-RS1              & 69.8(0.3)               & 25.2(1.1)              & 19.9(0.7)             & 66.0(1.3)             & 21.2(0.4)             & 40.4(0.4) \\
      Open-RS2              & 69.7(0.2)               & 23.3(0.9)              & 17.2(0.8)             & 64.8(1.1)             & 21.9(0.3)             & 39.4(0.3) \\
      Open-RS3              & 69.5(0.3)               & 24.8(0.9)              & 21.2(0.9)             & 64.4(0.7)             & 22.3(0.3)             & 40.4(0.3) \\
      Tina                  & 69.57(0.19)             & 25.2(1.2)              & 19.0(0.8)             & 63.4(0.8)             & 21.9(0.3)             & 39.8(0.3) \\
      \midrule
      \textsc{Still-3}-LoRA & \green{67}{72.1(0.3)}   & 24.3(1.3)              & 21.0(0.7)             & 67.3(1.1)             & 21.9(0.4)             & \green{29}{41.3(0.4)} \\
      DeepScaleR-LoRA       & 70.7(0.4)               & 23.6(1.0)              & 20.3(0.9)             & 66.7(1.2)             & 21.4(0.2)             & 40.5(0.4) \\
      \midrule
      \textsc{Mada}         & \green{63}{72.0(0.2)}   & \green{56}{26.2(0.7)}  & 20.0(0.9)             & \green{42}{68.7(1.2)} & 22.5(0.3)             & \green{41}{41.9(0.3)} \\
      \bottomrule
    \end{tabular}
    \begin{tablenotes}[flushleft]
      \footnotesize
      \setlength\labelsep{0pt}
    \item \emph{Note}: standard errors of the mean are reported throughout; cell colours indicate performance relative to the baseline DeepSeek-R1 model (red for worse, green for better), colour intensity scales with the magnitude of the difference, and white marks non-significant differences.
    \end{tablenotes}
  \end{threeparttable}
\end{table}

Aside from the two data-heavy baselines (DeepScaleR and \textsc{Still-3}), the \textsc{Mada} agents are the only ones to improve clearly and consistently over the base model with non-overlapping error bars, and the only ones to outperform it on every benchmark.

We state the gap plainly: \textsc{Mada} does not match DeepScaleR or \textsc{Still-3}.
Both are trained on substantially larger corpora with full-model fine-tuning\---DeepScaleR and \textsc{Still-3} update all \num{1.78e9} parameters, against \num{110e6} trainable parameters in our setup\---so the comparison most relevant to our claim is not raw accuracy but accuracy gain per unit of training cost, which we analyse in~\cref{sec:costs-performance}.
Notably, when DeepScaleR and \textsc{Still-3} are re-trained with LoRA on the same method and data (DeepScaleR-LoRA and \textsc{Still-3}-LoRA), their accuracy drops to \num{40.5} and \num{41.3} respectively: DeepScaleR-LoRA falls significantly below \textsc{Mada} ($p = \num{0.015}$), while \textsc{Still-3}-LoRA is statistically indistinguishable from it ($p = \num{0.280}$).
The full fine-tuning advantage of these stronger baselines therefore rests largely on data scale rather than on a mechanism that LoRA preserves, and at an equal trainable-parameter budget \textsc{Mada} matches or exceeds plain LoRA re-training.

\Cref{tab:significance} reports the statistical significance of \textsc{Mada} against every other model.
The \num{+2.0} point gain over the base model is significant ($p < \num{0.001}$), confirming that the improvement is not an artifact of decoding variance.
\textsc{Mada} also significantly outperforms every parameter-efficient community fine-tune (Open-RS1, Open-RS2, Open-RS3, and Tina), and no matched-budget method significantly exceeds it.
Consistent with our framing, it remains significantly below the two data-heavy, fully fine-tuned baselines (DeepScaleR and \textsc{Still-3}).
The full pairwise matrix of $p$-values is given in~\cref{sec:significance-full}.

\begin{table}[t]
  \centering
  \begin{threeparttable}
    \caption{\textsc{Mada} accuracy against each baseline}\label{tab:significance}
    \begin{tabular}{l S[table-format=2.1(2)] S[table-format=+1.1(2)] c S[table-format=<1.3]@{}l}
      \toprule
      Baseline              & {Accuracy}           & {Difference}         & {$\mathrm{CI}_{\qty{95}{\percent}}$} & \multicolumn{2}{c}{$p$} \\
      \midrule
      DeepSeek-R1           & \green{80}{39.9(3)}  & \green{80}{+2.0(5)}  & {$[+1.0, +3.0]$}                     & <0.001 & $^{***}$ \\
      \midrule
      \textsc{Still-3}      & \red{52}{43.1(4)}    & \red{52}{-1.3(5)}    & {$[-2.3, -0.2]$}                     & 0.020  & $^{*}$ \\
      DeepScaleR            & \red{96}{44.3(3)}    & \red{96}{-2.4(5)}    & {$[-3.4, -1.5]$}                     & <0.001 & $^{***}$ \\
      Open-RS1              & \green{60}{40.4(4)}  & \green{60}{+1.5(5)}  & {$[+0.4, +2.5]$}                     & 0.011  & $^{*}$ \\
      Open-RS2              & \green{100}{39.4(3)} & \green{100}{+2.5(5)} & {$[+1.5, +3.5]$}                     & <0.001 & $^{***}$ \\
      Open-RS3              & \green{56}{40.4(3)}  & \green{56}{+1.4(5)}  & {$[+0.5, +2.4]$}                     & 0.006  & $^{**}$ \\
      Tina                  & \green{84}{39.8(3)}  & \green{84}{+2.1(5)}  & {$[+1.0, +3.1]$}                     & <0.001 & $^{***}$ \\
      \midrule
      \textsc{Still-3}-LoRA & 41.3(4)              & +0.6(5)              & {$[-0.5, +1.7]$}                     & 0.280  & \\
      DeepScaleR-LoRA       & \green{56}{40.5(4)}  & \green{56}{+1.4(5)}  & {$[+0.3, +2.4]$}                     & 0.015  & $^{*}$ \\
      \midrule
      \textsc{Mada}         & 41.9(3)              & {---}                & {---}                                & {---} \\
      \bottomrule
    \end{tabular}
    \begin{tablenotes}[flushleft]
      \footnotesize
      \setlength\labelsep{0pt}
    \item \emph{Note}: standard errors of the mean are reported throughout;\enquote{accuracy} is the mean accuracy across all benchmarks; \enquote{difference} is the accuracy difference (\textsc{Mada} minus baseline) in points, shown with its \qty{95}{\percent} confidence interval; green favours \textsc{Mada}, red favours the baseline, colour intensity scales with the magnitude of the difference, and white marks non-significant differences; $p$-values are computed using Welch's two-sample $t$-test on per-seed average accuracy across the five benchmarks (ten seeds per model); $^{*} p < \num{0.05}$, \mbox{$^{**} p < \num{0.01}$, $^{***} p < \num{0.001}$}.
    \end{tablenotes}
  \end{threeparttable}
\end{table}

\subsection{Computational costs and performance}\label{sec:costs-performance}

To evaluate our method's efficiency, we assess the computational costs of \ac{mada}, considering both training-time and inference-time expenses.

\begin{figure}[h]
  \centering
  \begin{tikzpicture}
    \begin{axis}[
        grid=major,
        xlabel={Trainable parameters},
        xmin=0,
        xtick style={draw=none},
        ylabel={Accuracy},
        ymax=45,
        ymin=38.5,
        ytick={38, ..., 45},
        ytick style={draw=none},
      ]
      \addplot[mark=*] coordinates {
        (0, 39.9)
        (1.10e8, 41.9)
        (1.78e9, 44.3)
      };
      \addplot[
        mark=o,
        color=my_darkgray,
        only marks,
      ] coordinates {
        (1.78e9, 43.1)
        (1.78e9, 40.4)
        (1.78e9, 39.4)
        (1.78e9, 40.4)
        (3.69e7, 39.8)
        (1.48e8, 40.5)
        (1.48e8, 41.3)
      };
      \addplot[mark=*] coordinates {(1.10e8, 41.9)};
      \addplot[
        every mark/.append style={line width=1.5pt},
        only marks,
        mark=star,
        mark size=5pt,
      ] coordinates {(1.10e8, 41.9)};
      \node at (axis cs: 0, 39.9) [right, yshift=3pt] {DeepSeek-R1};
      \node at (axis cs: 1.10e8, 41.9) [below right] {\textbf{\textsc{Mada}}};
      \node at (axis cs: 1.48e8, 40.5) [right] {\textlight{DeepScaleR-LoRA}};
      \node at (axis cs: 1.48e8, 41.3) [below right] {\textlight{\textsc{Still-3}-LoRA}};
      \node at (axis cs: 1.78e9, 39.4) [below left] {\textlight{Open-RS2}};
      \node at (axis cs: 1.78e9, 40.4) [above left] {\textlight{Open-RS1}};
      \node at (axis cs: 1.78e9, 40.4) [below left] {\textlight{Open-RS3}};
      \node at (axis cs: 1.78e9, 43.1) [above left] {\textlight{\textsc{Still-3}}};
      \node at (axis cs: 1.78e9, 44.3) [above left] {DeepScaleR};
      \node at (axis cs: 3.69e7, 39.8) [below right] {\textlight{Tina}};
    \end{axis}
  \end{tikzpicture}
  \caption{
    Trainable parameters and accuracy.
    Darker points connected with a solid line represent the Pareto front.
    The star highlights our method, which achieves the highest gain per trainable parameter.
  }\label{fig:costs-performance}
\end{figure}
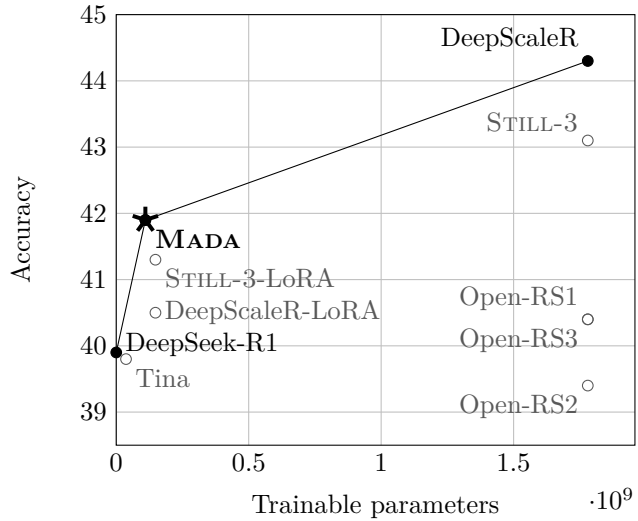

\paragraph{Training costs.}
To estimate training costs, we consider the number of trainable parameters per model.
The base model, DeepSeek-R1-Distill-Qwen-1.5B, contains approximately \num{1.78e9} parameters in total, and fully fine-tuned baselines update all of these.
In contrast, LoRA adapters add only \num{36.9e6} trainable parameters in the Tina model, \num{1.48e8} in DeepScaleR-LoRA and \textsc{Still-3}-LoRA, and \num{18.5e6} per agent in our setup, totaling \num{110e6}.
\Cref{fig:costs-performance,tab:costs-performance} report trainable parameters and performance gains, showing that our method achieves the largest improvement relative to trainable parameters.
While DeepScaleR achieves roughly twice the accuracy gain, it requires 16 times more trainable parameters, underscoring the advantage of \ac{mada} in enhancing reasoning performance under constrained training budgets.
In addition, DeepScaleR-LoRA and \textsc{Still-3}-LoRA have more trainable parameters than our setup yet perform worse, highlighting the training-parameter efficiency of our approach.

\begin{table}[t]
  \centering
  \begin{threeparttable}
    \caption{Computational costs and performance}\label{tab:costs-performance}
    \begin{tabular}{l S[table-format=1.2e2] S[table-format=+3] S[table-format=5] S[table-format=2.1]}
      \toprule
      Model                 & {Trainable parameters} & {Gain/parameter} & {Tokens/question} & {Critic improv.\ rate} \\
      \midrule
      DeepSeek-R1           & {\text{---}}           & {\text{---}}     & \gray{56}{30221}  & 18.5 \\
      \midrule
      \textsc{Still-3}      & \gray{100}{1.78e9}     & \green{10}{+18}  & \gray{13}{26629}  & \red{63}{17.5} \\
      DeepScaleR            & \gray{100}{1.78e9}     & \green{14}{+25}  & \gray{0}{25522}   & \red{100}{16.9} \\
      Open-RS1              & \gray{100}{1.78e9}     & \green{2}{+3}    & \gray{53}{29919}  & \red{19}{18.2} \\
      Open-RS2              & \gray{100}{1.78e9}     & \red{2}{-3}      & \gray{61}{30572}  & \green{50}{19.3} \\
      Open-RS3              & \gray{100}{1.78e9}     & \green{2}{+3}    & \gray{56}{30228}  & \red{25}{18.1} \\
      Tina                  & \gray{2}{3.69e7}       & \red{9}{-17}     & \gray{23}{27463}  & \green{6}{18.6} \\
      \midrule
      DeepScaleR-LoRA       & \gray{8}{1.48e8}       & \green{23}{+42}  & \gray{63}{30763}  & \red{13}{18.3} \\
      \textsc{Still-3}-LoRA & \gray{8}{1.48e8}       & \green{53}{+96}  & \gray{100}{33865} & \green{19}{18.8} \\
      \midrule
      \textsc{Mada}         & \gray{6}{1.10e8}       & \green{92}{+182} & \gray{99}{33818}  & \green{69}{19.6} \\
      \bottomrule
    \end{tabular}
    \begin{tablenotes}[flushleft]
      \footnotesize
      \setlength\labelsep{0pt}
    \item \emph{Note}: \enquote{gain/parameter} is the accuracy improvement over the base model divided by the number of trainable parameters (scaled by \num{e12}), a proxy for training-cost efficiency; \enquote{tokens/question} is the mean number of tokens generated per question under the full three-generator, two-round protocol, a proxy for inference cost; \enquote{critic improv.\ rate} is the percentage of questions on which the critics' final answer corrects the generators' consensus; relative to the base DeepSeek-R1 model, green cells indicate better performance and red cells indicate worse performance, with intensity scaling with the magnitude of the difference; grey encodes magnitude only, where better or worse does not apply.
    \end{tablenotes}
  \end{threeparttable}
\end{table}

\paragraph{Inference costs.}
The training-parameter savings come at a real inference-time price, which we report in~\cref{tab:costs-performance}.
The protocol issues six agent calls per question (three generators, then three critics) across two rounds, and \textsc{Mada} generates \num{33818} tokens per question\---among the highest of any model and roughly an order of magnitude more than a single forward pass.
For latency-sensitive deployment this is the method's main practical limitation, and any efficiency claim concerns training parameters, not inference compute.
Yet the token count alone misses a structural benefit: \textsc{Mada} achieves the highest critic improvement rate (\qty{19.6}{\percent}), correcting wrong generator answers more often than any other model\---a direct consequence of the counterfactual advantage, which rewards critics for being correct when the generators are not.
Baselines that place standard single-agent models in the critic role show lower improvement rates even when they generate comparable numbers of tokens, indicating that \textsc{Mada}'s gains stem from the critic's learned corrective behaviour rather than from deliberation volume alone.

\subsection{Ablation studies}\label{sec:ablations}

To assess the effectiveness of our design choices, we conducted several ablation studies, summarized in~\cref{tab:ablations}.

\begin{table}[t]
  \centering
  \begin{threeparttable}
    \caption{Ablation studies}\label{tab:ablations}
    \begin{tabular}{l S[table-format=2.1(2)] S[table-format=+1.1(2)] c S[table-format=<1.3]@{}l}
      \toprule
      Setting                  & {Accuracy}            & {Difference}          & $\mathrm{CI}_{\qty{95}{\percent}}$ & \multicolumn{2}{c}{$p$} \\
      \midrule
      2 rounds                 &                       & \\
      3 generators + 3 critics & 41.9(0.3)             & \text{---}            & \text{---}                         & \text{---} & \\
      Counterfactual           &                       & \\
      \midrule
      1 round                  & \red{100}{36.7(0.3)}  & \red{100}{-5.2(0.5)}  & $[-6.2, -4.2]$                     & <0.001     & $^{***}$ \\
      3 rounds                 & \green{27}{43.3(0.4)} & \green{27}{+1.4(0.5)} & $[+0.3, +2.5]$                     & 0.016      & $^{*}$ \\
      \midrule
      1 generator + 1 critic   & \red{50}{39.3(0.3)}   & \red{50}{-2.6(0.5)}   & $[-3.6, -1.6]$                     & <0.001     & $^{***}$ \\
      2 generators + 2 critics & \red{17}{41.0(0.2)}   & \red{17}{-0.9(0.4)}   & $[-1.8, \phantom{-}0.0]$           & 0.050      & $^{*}$ \\
      \midrule
      Homogeneous              & 41.3(0.2)             & -0.6(0.4)             & $[-1.5, +0.3]$                     & 0.166      & \\
      \midrule
      No counterfactual        & 41.1(0.3)             & -0.8(0.5)             & $[-1.8, +0.2]$                     & 0.095      & \\
      \bottomrule
    \end{tabular}
    \begin{tablenotes}[flushleft]
      \footnotesize
      \setlength\labelsep{0pt}
    \item \emph{Note}: standard errors of the mean are reported throughout; \enquote{difference} is the mean accuracy difference (main experiment minus ablation setting) in points, shown with its \qty{95}{\percent} confidence interval; green favours the ablation setting, red favours the main experiment, colour intensity scales with the magnitude of the difference, and white marks non-significant differences; $p$-values are computed using Welch's two-sample $t$-test on per-seed average accuracy across the five benchmarks (ten seeds per model); $^{*} p < \num{0.05}$, $^{**} p < \num{0.01}$, $^{***} p < \num{0.001}$.
    \end{tablenotes}
  \end{threeparttable}
\end{table}

\paragraph{Number of rounds.}
We compared the accuracy of the main experiment with both one-round and three-round debate variants.
A single round is effectively equivalent to generating multiple answers independently and evaluating them without any interaction.
As expected, this results in significantly worse performance compared to the standard two-round setup.
Adding a third round yields a further improvement, although the gain is smaller.
This suggests that iterative interaction between agents is an important driver of performance, while additional rounds provide diminishing returns.

\paragraph{Number of agents.}
We compared the main configuration with smaller debate ensembles consisting of one and two generator--critic pairs. 
The three-agent configuration outperforms smaller ensembles, suggesting that further scaling could yield additional gains; however, adding agents would require retraining the full pipeline from scratch due to the disjoint data partitioning scheme, making this a non-trivial extension that we leave for future work.

\paragraph{Agent diversity.}
To assess the role of diversity, we evaluated a homogeneous configuration in which all generators share a single generator checkpoint and all critics share a single critic checkpoint, as opposed to the three independently trained checkpoints used in the main experiment.
This configuration is \num{0.6} points below the main experiment, but the difference does not reach significance ($p = \num{0.166}$; $\mathrm{CI}_{\qty{95}{\percent}} = [\num{-1.5}, \num{+0.3}]$, which includes zero).
We therefore read agent diversity as a plausibly helpful but unconfirmed factor at this sample size: the point estimate favours diverse checkpoints, yet ten seeds do not provide the power to distinguish it from no effect.

\paragraph{Advantage computation.}
We trained an additional set of critics using the standard \ac{grpo} advantage, removing the counterfactual component introduced in~\cref{sec:training-critics}.
This variant is \num{0.8} points below the full \ac{mada} configuration; the drop is consistent in direction but does not reach significance at ten seeds ($p = \num{0.095}$; $\mathrm{CI}_{\qty{95}{\percent}} = [\num{-1.8}, \num{+0.2}]$).
We therefore do not rest the case for the counterfactual advantage on this ablation alone.
Its more direct signature is the critic improvement rate analysed in~\cref{sec:costs-performance}: the counterfactual-trained critics correct wrong generator answers more often (\qty{19.6}{\percent}) than any baseline placed in the critic role, a mechanism-level effect that does not depend on this accuracy difference clearing a significance threshold.
Read together, the directional accuracy drop and the corrective-behaviour evidence support the counterfactual component, while its marginal contribution to end-task accuracy remains within the noise at this sample size.

\paragraph{Debate contribution.}
\Cref{tab:single-agent} shows the models' performance in a single-agent setting, without any debate mechanism.
As expected, all models experience a drop in performance when debate is removed.
However, the drop is largest for \textsc{Mada} (\num{-5.2} points): this indicates that \textsc{Mada} relies more strongly on the debate process than the baseline models, suggesting that the training procedure encourages specialization of generator and critic roles rather than simply improving single-agent reasoning.

\begin{table}[t]
  \centering
  \begin{threeparttable}
    \caption{Single-agent performance}\label{tab:single-agent}
    \begin{tabular}{l S[table-format=2.1(2)] S[table-format=+1.1(2)] c S[table-format=<1.3]@{}l}
      \toprule
      Model                 & {Single agent}       & {Difference}        & $\mathrm{CI}_{\qty{95}{\percent}}$ & \multicolumn{2}{c}{$p$} \\
      \midrule
      DeepSeek-R1           & 36.1(3)              & \gray{6}{-3.8(5)}   & $[-4.7, -2.8]$                     & <0.001 & $^{***}$ \\
      \midrule
      \textsc{Still-3}      & \green{100}{39.5(4)} & \gray{0}{-3.7(5)}   & $[-4.7, -2.6]$                     & <0.001 & $^{***}$ \\
      DeepScaleR            & \green{94}{39.3(3)}  & \gray{89}{-5.0(4)}  & $[-5.9, -2.4]$                     & <0.001 & $^{***}$ \\
      Open-RS1              & 35.9(4)              & \gray{57}{-4.6(5)}  & $[-5.7, -3.4]$                     & <0.001 & $^{***}$ \\
      Open-RS2              & 35.3(3)              & \gray{30}{-4.1(5)}  & $[-5.1, -3.1]$                     & <0.001 & $^{***}$ \\
      Open-RS3              & 36.2(3)              & \gray{37}{-4.2(4)}  & $[-5.2, -3.3]$                     & <0.001 & $^{***}$ \\
      Tina                  & 35.2(3)              & \gray{64}{-4.7(5)}  & $[-5.7, -3.7]$                     & <0.001 & $^{***}$ \\
      \midrule
      DeepScaleR-LoRA       & 36.9(4)              & \gray{25}{-3.6(5)}  & $[-4.7, -2.5]$                     & <0.001 & $^{***}$ \\
      \textsc{Still-3}-LoRA & 36.3(4)              & \gray{96}{-5.0(5)}  & $[-6.2, -3.9]$                     & <0.001 & $^{***}$ \\
      \midrule
      \textsc{Mada}         & 36.7(3)              & \gray{100}{-5.2(5)} & $[-6.2, -4.2]$                     & <0.001 & $^{***}$ \\
      \bottomrule
    \end{tabular}
    \begin{tablenotes}[flushleft]
      \footnotesize
      \setlength\labelsep{0pt}
    \item \emph{Note}: standard errors of the mean are reported throughout; for \textsc{Mada}, the score is the average performance of the fine-tuned generator agents; relative to the base DeepSeek-R1 model, green cells indicate better performance, intensity scales with the magnitude of the difference, and white marks non-significant differences; \enquote{difference} is the mean accuracy difference (full debate minus single-agent setting) in points; grey encodes magnitude only, where better or worse does not apply; $p$-values are computed using Welch's two-sample $t$-test on per-seed average accuracy across the five benchmarks (ten seeds per model); $^{*} p < \num{0.05}$, $^{**} p < \num{0.01}$, $^{***} p < \num{0.001}$.
    \end{tablenotes}
  \end{threeparttable}
\end{table}

\paragraph{Ablation takeaways.}
The debate structure is the dominant and statistically robust driver: removing rounds or shrinking the ensemble produces large, significant drops.
Agent diversity and the counterfactual advantage contribute smaller, directionally consistent effects that do not individually reach significance at ten seeds; for the counterfactual component the accuracy ablation is corroborated by its distinct mechanism-level signature, the highest critic improvement rate among all models.
We thus frame the debate structure as established and the training-objective refinements as supported but more lightly powered.

\section{Related work}\label{sec:related-work}

\paragraph{Reinforcement learning for LLM reasoning.}
\Ac{rl} has become a key tool for enhancing reasoning in language models, particularly in low-resource settings.
\Ac{grpo} \citep{shao2024deepseekmath} removes the value model and uses group-normalized rewards for stability in long-form tasks, while \ac{dapo} \citep{yu2025dapo} adds dynamic sampling and structured reward shaping.
We draw on both, but unlike \textsc{dapo} or value-model-based token-level methods \citep{yue2025vapo}, we derive answer-level correctness and counterfactual signals from multi-agent debate, yielding a lighter, fully interpretable reward~pipeline.

\paragraph{Multi-agent debate and adjudication.}
Composing several model instances and letting them deliberate is an established route to better reasoning.
\citet{du2024improving} have multiple instances propose and revise answers over rounds to converge on a consensus, while \citet{liang2024encouraging} cast the interaction as an adversarial debate arbitrated by a judge, and \citet{irving2018ai} motivate debate as a mechanism for scalable oversight.
A parallel line aggregates rather than argues\---self-consistency \citep{wang2022self} by majority vote, \ac{llm}-as-judge by a separate adjudicator\---but these methods are training-free: they improve a fixed model at inference time without adapting the agents.
Closer to us, \citet{park2025maporl} proposed \ac{maporl}, which guides \ac{rl} with a learned verifier, and \citet{subramaniam2025multiagent} use debate to generate data for supervised fine-tuning.
We differ in what we train: rather than learning a verifier or distilling transcripts, we fold one debate-derived statistic\---the generator ensemble's per-instance accuracy\---directly into the critic's advantage, so critics are optimized online to correct consensus errors without any auxiliary model.

\paragraph{Compact models and efficient multi-agent RL.}
Other approaches explore multi-agent \ac{rl}, such as \acl{marft} (\acs{marft}; \citealp{liao2025marft}), which jointly trains agents with shared gradients but adds coordination overhead.
We instead keep agents independent and assign credit locally via debate, enabling efficient \ac{lora} training \citep{hu2022lora}.
Our work also connects to compact-model reasoning: \acl{tina} (\acs{tina}; \citealp{wang2025tina}) shows small \acp{llm} are competitive with \ac{lora} and minimal \ac{rl}, and \acl{oreo} (\acs{oreo}; \citealp{wang2025offline}) uses offline \ac{rl} with learned critics; unlike these, we train online with interpretable, multi-agent feedback.

\section{Conclusion}\label{sec:conclusion}

We introduce \ac{mada}, a parameter-efficient post-training framework that specializes a compact \ac{llm} into generator and critic roles and trains the critics with a counterfactual advantage that rewards them for correcting generator consensus.
Using only \ac{lora} adapters and \ac{grpo}, it improves a \qty{1.5}{B} model by \num{2.0} points across five mathematical reasoning benchmarks while updating \num{16} times fewer parameters than fully fine-tuned baselines\---the largest accuracy gain per trainable parameter of any model we evaluate.
It approaches but does not surpass the strongest data-heavy baselines, whose matched-parameter \ac{lora} re-trainings do not exceed \ac{mada}, indicating that their advantage rests on data scale rather than on a component \ac{lora} retains.
A controlled analysis ties these gains to learned corrective behaviour rather than deliberation volume, positioning \ac{mada} as a practical recipe for reasoning gains on compact models under tight training budgets.

Several limitations remain and sharpen the agenda.
First, the most informative comparison is still open: we have not measured \ac{mada} against training-free debate, voting, or \ac{llm}-as-judge baselines at matched inference budget, so we cannot yet claim it beats spending the same test-time compute on a strong single model.
Second, we characterize the counterfactual advantage empirically but do not analyse its convergence or effect on gradient variance relative to standard \ac{grpo}.
Third, our Welch tests use per-seed average accuracy (\cref{tab:significance}); per-benchmark and paired tests, bootstrap intervals, and multiple-comparison correction remain future work.
Fourth, we evaluate only a single \qty{1.5}{B} model on mathematical benchmarks; scaling toward \qty{4}{B} and transfer to scientific reasoning or planning are untested.
Finally, the multi-round protocol raises inference latency (\cref{sec:costs-performance}), which distilling the debate into a single model could mitigate.
We will release our code, trained \ac{lora} adapters, and evaluation scripts.




\bibliography{bib}
\bibliographystyle{tmlr}

\appendix

\section{Technical appendix}

\subsection{Prompts}

This section reports the various prompts used for text generation.

\paragraph{Math reasoning.}
This is the text that is prepended to all queries to the models:
\begin{prompt}
Solve the following math problem efficiently and clearly. The last line of your response should be of the following format: 'Therefore, the final answer is: $\\boxed{{ANSWER}}$. I hope it is correct' (without quotes) where ANSWER is just the final number or expression that solves the problem. Think step by step before answering.

{question}
\end{prompt}

\paragraph{Multi-agent debate.}
When concatenating outputs during the multi-agent debate, this was inserted first:
\begin{prompt}
These are the recent opinions from other agents:
\end{prompt}
followed by this text for each agent:
\begin{prompt}
One agent's response:
```
{response}
```
\end{prompt}
and ending with:
\begin{prompt}
Using each response as additional advice, can you give an updated answer to the question?
\end{prompt}

\FloatBarrier{}

\subsection{Hyperparameters}

Training hyperparameters are detailed in~\cref{tab:training-hyperparameters}.
These values were chosen based on empirical validation and informed by the tuning strategies proposed by previous works.

\begin{table}[t]
  \caption{Training hyperparameters}\label{tab:training-hyperparameters}
  \centering
  \begin{tabular}{l S[table-alignment-mode=none]}
    \toprule
    Batch size                     & 16 \\
    Learning rate                  & 1e-06 \\
    Warmup ratio                   & 0.1 \\
    \midrule
    AdamW weight decay             & 0 \\
    AdamW $\beta_1$                & 0.9 \\
    AdamW $\beta_2$                & 0.999 \\
    AdamW $\varepsilon$            & 1e-8 \\
    \midrule
    \Ac{grpo} group size           & 4 \\
    \Ac{grpo} sampling temperature & 0.7 \\
    \Ac{grpo} $\beta$              & 0.04 \\
    \Ac{grpo} $\varepsilon$        & 0.2 \\
    \midrule
    \Ac{lora} rank                 & 16 \\
    \Ac{lora} $\alpha$             & 128 \\
    \Ac{lora} dropout              & 0.05 \\
    \bottomrule
  \end{tabular}
\end{table}

\Cref{tab:inference-hyperparameters} reports the hyperparameters used for text generation at inference.

\begin{table}[t]
  \caption{Inference hyperparameters}\label{tab:inference-hyperparameters}
  \centering
  \begin{tabular}{lS[table-alignment-mode=none]}
    \toprule
    Temperature              & 1.0 \\
    Top $p$                  & 1.0 \\
    Maximum generated tokens & 32768 \\
    \midrule
    Number of generators     & 3 \\
    Number of critics        & 3 \\
    Number of rounds         & 2 \\
    \midrule
    Samples per experiment   & 10 \\
    \bottomrule
  \end{tabular}
\end{table}

\subsection{Computing infrastructure}

Experiments were run on the Red Hat Enterprise Linux 9.5 operating system with \qty{32}{GB} of RAM\@.
Both training and inference were run on a single H200 GPU with \qty{141}{GB} of video RAM\@.

\subsection{Full pairwise significance}\label{sec:significance-full}

\Cref{fig:accuracy-full} reports all pairwise mean accuracy differences (with standard errors) as an effect-size matrix, and~\cref{tab:significance-full} the corresponding Welch $t$-test $p$-values.
With ten models there are \num{45} pairwise comparisons and we apply no multiple-comparison correction, so we treat these as exploratory; the pre-specified comparisons of \textsc{Mada} against each baseline in~\cref{tab:significance} carry the formal claims.

\begin{sidewaystable}[p]
  \centering
  \begin{minipage}[t][0.35\textwidth][s]{\textheight}
    \centering
    \begin{threeparttable}
      \caption{Pairwise mean accuracy differences between models}\label{fig:accuracy-full}
      \sisetup{table-format=+1.1(2)}
      \begin{tabular}{l *{8}{S[table-column-width=\sigcolw, table-format=+1.1(2)]} S S[table-column-width=\sigcolw, table-format=+1.1(2)]}
        \toprule
                         & {DS-R1}             & {\textsc{Still-3}}  & {DSR}              & {Open-RS1}          & {Open-RS2}           & {Open-RS3}          & {Tina}              & {S3-LoRA}           & {DSR-LoRA}          & {\textsc{Mada}} \\
        \midrule
        DS-R1            & {---}               & \red{67}{-3.3(5)}   & \red{90}{-4.4(4)}  & -0.5(5)             & +0.5(5)              & -0.6(4)             & +0.1(5)             & \red{29}{-1.4(5)}   & -0.6(5)             & \red{41}{-2.0(5)} \\
        \textsc{Still-3} & \green{67}{+3.3(5)} & {---}               & \red{22}{-1.1(5)}  & \green{55}{+2.7(5)} & \green{78}{+3.8(5)}  & \green{55}{+2.7(5)} & \green{67}{+3.3(5)} & \green{37}{+1.8(5)} & \green{53}{+2.6(5)} & \green{27}{+1.3(5)} \\
        DSR              & \green{90}{+4.4(4)} & \green{22}{+1.1(5)} & {---}              & \green{80}{+3.9(5)} & \green{100}{+4.9(4)} & \green{80}{+3.9(4)} & \green{92}{+4.5(4)} & \green{61}{+3.0(5)} & \green{78}{+3.8(5)} & \green{49}{+2.4(5)} \\
        Open-RS1         & +0.5(5)             & \red{55}{-2.7(5)}   & \red{80}{-3.9(5)}  & {---}               & +1.0(5)              & 0.0(5)              & +0.6(5)             & -0.9(5)             & -0.1(5)             & \red{31}{-1.5(5)} \\
        Open-RS2         & -0.5(5)             & \red{78}{-3.8(5)}   & \red{100}{-4.9(4)} & -1.0(5)             & {---}                & \red{22}{-1.1(5)}   & -0.4(5)             & \red{39}{-1.9(5)}   & \red{22}{-1.1(5)}   & \red{51}{-2.5(5)} \\
        Open-RS3         & +0.6(4)             & \red{55}{-2.7(5)}   & \red{80}{-3.9(4)}  & 0.0(5)              & \green{22}{+1.1(5)}  & {---}               & +0.6(5)             & -0.9(5)             & -0.1(5)             & \red{29}{-1.4(5)} \\
        Tina             & -0.1(5)             & \red{67}{-3.3(5)}   & \red{92}{-4.5(4)}  & -0.6(5)             & +0.4(5)              & -0.6(5)             & {---}               & \red{31}{-1.5(5)}   & -0.7(5)             & \red{43}{-2.1(5)} \\
        S3-LoRA          & \green{29}{+1.4(5)} & \red{37}{-1.8(5)}   & \red{61}{-3.0(5)}  & +0.9(5)             & \green{39}{+1.9(5)}  & +0.9(5)             & \green{31}{+1.5(5)} & {---}               & +0.8(5)             & -0.6(5) \\
        DSR-LoRA         & +0.6(5)             & \red{53}{-2.6(5)}   & \red{78}{-3.8(5)}  & +0.1(5)             & \green{22}{+1.1(5)}  & +0.1(5)             & +0.7(5)             & -0.8(5)             & {---}               & \red{29}{-1.4(5)} \\
        \textsc{Mada}    & \green{41}{+2.0(5)} & \red{27}{-1.3(5)}   & \red{49}{-2.4(5)}  & \green{31}{+1.5(5)} & \green{51}{+2.5(5)}  & \green{29}{+1.4(5)} & \green{43}{+2.1(5)} & +0.6(5)             & \green{29}{+1.4(5)} & {---} \\
        \bottomrule
      \end{tabular}
      \begin{tablenotes}[flushleft]
        \footnotesize
        \setlength\labelsep{0pt}
      \item \emph{Note}: each cell reports the mean accuracy difference (row model minus column model) in points, with its standard error of the mean; green indicates the row model is significantly more accurate and red significantly less accurate (Welch's two-sample $t$-test, $p<\num{0.05}$), with intensity scaling with the magnitude of the difference, while white marks non-significant pairs; \enquote{DS-R1} stands for DeepSeek-R1, \enquote{DSR} for DeepScaleR, and \enquote{S3} for \textsc{Still-3}.
      \end{tablenotes}
    \end{threeparttable}
    \vfill
    \begin{threeparttable}
      \caption{Pairwise statistical significance of mean accuracy differences between models}\label{tab:significance-full}
      \sisetup{table-format=<1.3}
      \begin{tabular}{l *{8}{S[table-column-width=\sigcolw]} S S[table-column-width=\sigcolw]}
        \toprule
                         & {DS-R1}            & {\textsc{Still-3}} & {DSR}              & {Open-RS1}         & {Open-RS2}         & {Open-RS3}         & {Tina}             & {S3-LoRA}          & {DSR-LoRA}         & {\textsc{Mada}} \\
        \midrule
        DS-R1            & {---}              & \gray{100}{<0.001} & \gray{100}{<0.001} & 0.296              & 0.293              & 0.233              & 0.895              & \gray{33}{0.011}   & 0.217              & \gray{100}{<0.001} \\
        \textsc{Still-3} & \gray{100}{<0.001} & {---}              & \gray{33}{0.025}   & \gray{100}{<0.001} & \gray{100}{<0.001} & \gray{100}{<0.001} & \gray{100}{<0.001} & \gray{67}{0.003}   & \gray{100}{<0.001} & \gray{33}{0.020} \\
        DSR              & \gray{100}{<0.001} & \gray{33}{0.025}   & {---}              & \gray{100}{<0.001} & \gray{100}{<0.001} & \gray{100}{<0.001} & \gray{100}{<0.001} & \gray{100}{<0.001} & \gray{100}{<0.001} & \gray{100}{<0.001} \\
        Open-RS1         & 0.296              & \gray{100}{<0.001} & \gray{100}{<0.001} & {---}              & 0.055              & 0.973              & 0.254              & 0.115              & 0.863              & \gray{33}{0.011} \\
        Open-RS2         & 0.293              & \gray{100}{<0.001} & \gray{100}{<0.001} & 0.055              & {---}              & \gray{33}{0.033}   & 0.365              & \gray{67}{0.001}   & \gray{33}{0.037}   & \gray{100}{<0.001} \\
        Open-RS3         & 0.233              & \gray{100}{<0.001} & \gray{100}{<0.001} & 0.973              & \gray{33}{0.033}   & {---}              & 0.197              & 0.093              & 0.877              & \gray{67}{0.006} \\
        Tina             & 0.895              & \gray{100}{<0.001} & \gray{100}{<0.001} & 0.254              & 0.365              & 0.197              & {---}              & \gray{67}{0.009}   & 0.185              & \gray{100}{<0.001} \\
        S3-LoRA          & \gray{33}{0.011}   & \gray{67}{0.003}   & \gray{100}{<0.001} & 0.115              & \gray{67}{0.001}   & 0.093              & \gray{67}{0.009}   & {---}              & 0.151              & 0.280 \\
        DSR-LoRA         & 0.217              & \gray{100}{<0.001} & \gray{100}{<0.001} & 0.863              & \gray{33}{0.037}   & 0.877              & 0.185              & 0.151              & {---}              & \gray{33}{0.015} \\
        \textsc{Mada}    & \gray{100}{<0.001} & \gray{33}{0.020}   & \gray{100}{<0.001} & \gray{33}{0.011}   & \gray{100}{<0.001} & \gray{67}{0.006}   & \gray{100}{<0.001} & 0.280              & \gray{33}{0.015}   & {---} \\
        \bottomrule
      \end{tabular}
      \begin{tablenotes}[flushleft]
        \footnotesize
        \setlength\labelsep{0pt}
      \item \emph{Note}: each cell reports the two-sided $p$-value of the Welch two-sample $t$-test for the difference between the row and column models (per-seed average accuracy, ten seeds per model); grey intensity reflects the significance level ($p < \num{0.05}$, $p < \num{0.01}$, $p < \num{0.001}$), and white indicates non-significance ($p \geq \num{0.05}$); \enquote{DS-R1} stands for DeepSeek-R1, \enquote{DSR} for DeepScaleR, and \enquote{S3} for \textsc{Still-3}.
      \end{tablenotes}
    \end{threeparttable}
  \end{minipage}
\end{sidewaystable}

\end{document}